# Vision-Language Models for Autonomous Driving: CLIP-Based Dynamic Scene Understanding

Mohammed Elhenawy, Huthaifa I. Ashqar, Andry Rakotonirainy, Taqwa I. Alhadidi, Ahmed Jaber, and Mohammad Abu Tami

*Abstract*—Scene understanding is essential for enhancing driver safety, generating human-centric explanations for Automated Vehicle (AV) decisions, and leveraging Artificial Intelligence (AI) for retrospective driving video analysis. This study developed a dynamic scene retrieval system using Contrastive Language–Image Pretraining (CLIP) models, which can be optimized for real-time deployment on edge devices. The proposed system outperforms state-of-the-art in-context learning methods, including the zero-shot capabilities of GPT-4o, particularly in complex scenarios. By conducting frame-level analysis on the Honda Scenes Dataset, which contains a collection of about 80 hours of annotated driving videos capturing diverse real-world road and weather conditions, our study highlights the robustness of CLIP models in learning visual concepts from natural language supervision. Results also showed that fine-tuning the CLIP models, such as ViT-L/14 and ViT-B/32, significantly improved scene classification, achieving a top F1-score of 91.1%. These results demonstrate the ability of the system to deliver rapid and precise scene recognition, which can be used to meet the critical requirements of Advanced Driver Assistance Systems (ADAS). This study shows the potential of CLIP models to provide scalable and efficient frameworks for dynamic scene understanding and classification. Furthermore, this work lays the groundwork for advanced autonomous vehicle technologies by fostering a deeper understanding of driver behavior, road conditions, and safety-critical scenarios, marking a significant step toward smarter, safer, and more context-aware autonomous driving systems.

## I. INTRODUCTION

UNDERSTANDING scenes is considered one of the most challenging tasks for improving driver safety and enhancing the performance of ADAS [1], [2]. The use of AI in this field has opened new dimensions for the retrospective analysis of driving videos. This enables the analysis of driver behavior in terms of gap identification in driving skills and assessment of overall driving performance [3], [4], [5]. The role of AI in improving scene understanding improving scene understanding is further emphasized by its ability to process massive volumes of driving data. For example, machine learning techniques, like Long Short-Term Memory (LSTM) networks, have been applied to classify and predict driver behavior based on real-time data [6].

It will have the possibility to develop systems able not only to learn a particular style of driving but also to adapt themselves to environmental conditions to optimize safety performance. Additionally, AI in ADAS means continuous monitoring of driver behavior, whereby at any given moment the system could give some feedback or even intervene [7]. AI significantly increases vehicular situational awareness and subsequently decision-making processes by processing and analyzing data from a wide range of sensors such as cameras and LiDAR [8]. Furthermore, driver acceptance of ADAS can be better understood by gaining insights into how drivers will respond to automated systems, especially in terms of reliance and trust [9]. The extent to which the drivers may be willing to utilize these systems can be greatly impacted by the driver's perception of the systems reliability [10], [11].

Hence it is vital to devise and build an ADAS which can support drivers and produce synergy between humans and machines. This necessity brings spotlight to systems that can understand and react to the ever-changing landscape of driving to improve overall safety and efficiency of driving. Driver training and understanding of these technologies are other important factors which also affect the working of the ADAS. Evidence suggests that training can impact drivers'

This work was funded partially by the Australian Government through the Australian Research Council Discovery Project DP220102598.
Corresponding author: Huthaifa I. Ashqar.

Mohammed Elhenawy is with CARRS-Q and Centre for Data Science, Queensland University of Technology, Kelvin Grove QLD 4059, Australia (e-mail: mohammed.elhenawy@qut.edu.au).

Huthaifa I. Ashqar is with the Data Science and AI Department, Arab American University, Jenin P.O Box 240, Palestine, and he is with Artificial Intelligence Program, Fu Foundation School of Engineering and Applied Science, Columbia University, New York, NY 10027, USA (e-mail: Huthaifa.ashqar@aaup.edu).

Andry Rakotonirainy is with CARRS-Q, Queensland University of Technology, Kelvin Grove QLD 4059, Australia (e-mail: r.andry@qut.edu.au).

Taqwa I. Alhadidi is with Civil Engineering Department, Al-Ahliyya Amman University, Amman 19328 e-mail: t.alhadidi@ammanu.edu.jo).

Ahmed Jaber is with Department of Transport Technology and Economics, Faculty of Transportation Engineering and Vehicle Engineering, Budapest University of Technology and Economics, Műegyetem rkp. 3, H-1111 Budapest, Hungary (e-mail: ahjaber6@edu.bme.hu).

Mohammad Abu Tami is with Natural, Engineering and Technology Sciences Department, Arab American University, Jenin P.O Box 240, Palestine (e-mail: m.abutami@student.aaup.edu).

knowledge regarding the capabilities and limitations of ADAS, and consequently, acceptance and trust in automation [12]. This is specifically relevant because motorists rely more and more on these systems for safety and assistance. Driver Mental Models: The design of the ADAS must therefore take into account not only the technological capacities of the drivers, but also their mental models and interaction with the systems in question [13]. This ability of AI to perceive the scene has been extended by its ability to scan millions of driving hours of data.

This study presents a new framework for dynamic scene understanding enabled by the CLIP model. The key contributions of this work are as follows: (1) We propose a CLIP-based retrieval system for real-time decision-making in an ADAS and demonstrate its effectiveness in understanding dynamic driving scenes, which can be used for real-time deployment on edge devices. (2) We benchmarked the performance of the CLIP model against state-of-the-art in-context learning methods, including the zero-shot capabilities of GPT-4o, highlighting the superior performance of CLIP for scene classification. (3) We apply a frame-level analysis of the publicly available Honda Scenes Dataset, incorporating diverse road and environmental conditions to enable the robust classification of traffic scenes. (4) The study then shows how multimodal AI can add depth to the recognition of road conditions and driver behavior to provide meaningful insights into autonomous driving systems and road safety research. Thus, this study lays the foundation for further research on the effective integration of language-based supervision with visual scene recognition within ADAS applications.

The rest of the paper is organized as follows: Section II describes the literature review on current approaches and advancements in the field of scene understanding and ADAS technologies. Section III provides a background and comparison between CLIP models used in this study. Section IV outlines the methodology, detailing how the proposed CLIP-based retrieval system was implemented. Section V discusses the dataset and preprocessing, focusing on the Honda Scenes Dataset and how it was prepared for analysis. The results can be found in Section VI, where the system performance is compared with the state-of-the-art methods. Section VII discusses the findings and their implications. Finally, the conclusions are drawn in Section VIII, summarizing the paper's contributions toward advances in multimodal AI for dynamic scene understanding.

II. LITERATURE REVIEW

Scene understanding is very significant in transportation, furthered by the need for the interaction of a vehicle with the road and its surroundings [14], [15]. In relation, the work by Pham et al. regarding joint geometric and object segmentation stipulates the significance of proper identification and classification of objects within indoor scenes into the settings of transportation [16]. The improvement in accuracy while detecting objects helps vehicles to make better interpretations and subsequent decisions.

These basic methods are all about the employment of convolutional neural networks for scene analysis and classification according to their semantic contents. CNNs are pretty good in extracting features from images, thus providing a way of identifying objects and their relations within a scene. For example, Li et al. have indicated that deep learning approaches have dominated the evolution of scene understanding, especially with CNNs that outperform the traditional approaches such as K-means and SVMs [17]. Ni et al. gave more weight to this claim when they proposed a deep network-based improved method of scene classification specifically designed for the self-driving car, enhanced by a novel Inception module in both global and local features extraction [18]. The robustness and accuracy of feature fusion tremendously upgrade the performance of scene classification for the safe driving of an ego vehicle in a complex real environment.

Complementary to CNNs, the integration of methods for 3D scene understanding will play an increasingly important role in tasks such as autonomous vehicles. The model proposed by Han et al. predicts a 3D scene graph that describes entities within a scene and gives semantic relationships among these entities for enhanced environmental understanding [19]. This approach bridges an important gap in the literature on scene understanding, because traditional methods are usually insensitive to the relations between objects. Moreover, LiDAR combined with deep learning has been demonstrating encouraging results on the estimation of scene flow and relative pose, which was introduced by Li et al. in the study of LiDAR odometry for unmanned ground vehicles [20]. Taken together, these technologies enable real-time processing and, therefore, for precise mapping in dynamic environments, which is very important for good navigation.

Another challenge was the proper detection and classification of various road conditions, from potholes to wet surfaces, through novel machine learning. Vernekar et al. developed another type of pipeline using pretrained models to analyze live images captured from traffic cameras. This refines the existing feature extraction process to enhance the accuracy in detection [21]. This approach not only enhances the reliability of understanding scenes, but also contributes significantly to the safety of transportation systems by enabling timely responses to hazardous situations.

The use of synthetic data generation to train AI models is increasing in the field of scene understanding. Holst et al. talk about generating synthetic training data for robotics applications, which can be adapted for various transportation scenarios [22]. Using simulations of diverse environments and conditions, researchers can create robust datasets that improve the performance of machine learning models in real-world applications.

Another critical aspect of scene understanding related to transportation is the integration of multimodal data sources. For instance, the work of Zipfl and Zöllner insists that image-based approaches must combine with graph-based methods representing spatial environmental factors and their relations regarding traffic participants [23].

This paper presents a new framework of dynamic understanding based on the CLIP model. The main contributions of this work are as follows: (1) A CLIP-based retrieval system for real-time decision-making in ADAS applications is proposed and demonstrates its effectiveness in understanding dynamic driving scenes. (2) CLIP performance is compared to the state-of-the-art in-context learning methods, including zero-shot performance for GPT-4o. This is in order to show that CLIP performs best on the task of scene classification. (3) Frame-level analysis is done on the publicly available Honda Scenes Dataset, which includes a wide variety of road and environmental conditions that can enable robust traffic scene classification. (4) The work investigates how to effectively integrate language-based supervision with visual scene recognition, providing meaningful insights into road safety research and autonomous driving systems by laying the bedrock for further advancements in multimodal AI.

## III. CONTRASTIVE LANGUAGE–IMAGE PRE-TRAINING (CLIP)

CLIP models have significantly advanced the field of vision-language understanding by learning visual representations from natural language supervision [24]. One advantage of using CLIP is that it can be deployed on edge devices with careful consideration of hardware capabilities, model size, and optimization techniques. While smaller variants (e.g., ViT-B/32) are more practical for edge use, larger variants like ViT-L/14 may require significant optimizations or hybrid deployment strategies. The choice ultimately depends on the application's real-time requirements, available resources, and the complexity of the scene understanding tasks. Various CLIP model architectures have been developed, each with distinct characteristics concerning model size, processing speed, VRAM requirements, and architectural design. This document compares five prominent CLIP models—ViT-B/32, ViT-B/16, ViT-L/14, RN50, and RN101—in terms of their number of parameters, processing speed, VRAM requirements for embedding, and architectural differences. The goal is to provide insights into their suitability for different applications, aiding in the selection of an appropriate model based on specific requirements. The following is a comparison between the different CLIP models.

### A. Number of Parameters

The number of parameters in a model is a crucial factor influencing its computational requirements and potential performance. Models with more parameters can capture more complex patterns but require more computational resources for training and inference. TABLE I summarizes the number of parameters for each CLIP model under consideration. The text encoder parameters are consistent across models except for ViT-L/14, which uses a larger text encoder.

TABLE I
NUMBER OF PARAMETERS IN CLIP MODELS

| Model | Image Encoder Parameters | Text Encoder Parameters | Total Parameters |
|---|---|---|---|
| CLIP ViT-B/32 | ~86 million | ~63 million | ~149 million |
| CLIP ViT-B/16 | ~86 million | ~63 million | ~149 million |
| CLIP ViT-L/14 | ~304 million | ~123 million | ~427 million |
| CLIP RN50 | ~102 million | ~63 million | ~165 million |
| CLIP RN101 | ~152 million | ~63 million | ~215 million |

### B. Processing Speed (Frames per Second)

Processing speed is a critical factor, especially for real-time applications. It depends on the model's complexity, computational requirements, and the hardware used. TABLE II provides approximate frames per second (FPS) for each model on a high-end GPU (e.g., NVIDIA RTX 3090). It is worth noting that these are rough estimates. Actual performance may vary based on hardware specifications, software optimizations, and batch size.

TABLE II
APPROXIMATE PROCESSING SPEED OF CLIP MODELS

| Model | Approximate FPS |
|---|---|
| CLIP ViT-B/32 | 150–200 |
| CLIP ViT-B/16 | 80–120 |
| CLIP ViT-L/14 | 30–60 |
| CLIP RN50 | 120–160 |
| CLIP RN101 | 70–100 |

### C. Models Architecture

The architecture of a model influences its ability to capture features, computational efficiency, and suitability for different tasks. The CLIP models considered here are based on two primary architectures: Vision Transformers (ViT) and ResNets.

Vision Transformers apply transformer architectures to sequences of image patches, leveraging self-attention mechanisms to model global relationships within an image [25]. Key architectural details of the ViT-based CLIP models are provided in TABLE III. Note that the number of patches is calculated by dividing the image dimensions by the patch size and squaring the result.

TABLE III
ARCHITECTURAL DETAILS OF ViT-BASED CLIP MODELS

| Model | Type | Layers | Hidden Size | Attention Heads | Patch Size | Number of Patches (224×224 image) |
|---|---|---|---|---|---|---|
| ViT-B/32 | ViT-Base | 12 | 768 | 12 | 32×32 | 49 |
| ViT-B/16 | ViT-Base | 12 | 768 | 12 | 16×16 | 196 |
| ViT-L/14 | ViT-Large | 24 | 1024 | 16 | 14×14 | 256 |

ResNet models utilize convolutional neural networks (CNN) with residual connections, enabling the training of deeper networks by mitigating the vanishing gradient problem [26]. Architectural details of the ResNet-based CLIP models are provided in TABLE IV.

TABLE IV
ARCHITECTURAL DETAILS OF RESNET-BASED CLIP MODELS

| Model | Type | Layers | Block Type |
|---|---|---|---|
| CLIP RN50 | ResNet-50 | 50 | Bottleneck blocks |
| CLIP RN101 | ResNet-101 | 101 | Bottleneck blocks |

However, ViT models process images as sequences of patches and rely on self-attention mechanisms to capture global context. In contrast, ResNet models use convolutional layers to capture local spatial hierarchies and employ residual connections to facilitate training of deep networks. Moreover, ViT-L/14 is significantly larger than the ViT-B models and ResNet models due to more layers and a larger hidden size. It is also worth noting that smaller patch sizes in ViT models (e.g., ViT-B/16 and ViT-L/14) result in more patches per image, allowing the model to capture finer details but increasing computational load.

### D. VRAM Requirements for Embedding

The amount of VRAM (Video Random Access Memory) required for embedding using each model is a crucial consideration, especially when deploying models on GPUs with limited memory. VRAM requirements can vary based on factors such as batch size and implementation details. Approximate requirements for inference with a batch size of 1 are provided in TABLE V. Note that VRAM requirements increase with larger batch sizes and higher-resolution images.

TABLE V
APPROXIMATE VRAM REQUIREMENTS FOR EMBEDDING

| Model | Approximate VRAM Requirement (GB) |
|---|---|
| CLIP ViT-B/32 | 1–2 |
| CLIP ViT-B/16 | 2–3 |
| CLIP ViT-L/14 | 4–6 |
| CLIP RN50 | 2–3 |
| CLIP RN101 | 3–4 |

TABLE V showed that CLIP ViT-B/32 requires the least VRAM among ViT models, making it suitable for devices with limited GPU memory. It also showed that CLIP ViT-L/14 has the highest VRAM requirement due to its large number of parameters and deeper architecture. Nonetheless, ResNet models generally have moderate VRAM requirements, with RN101 requiring more memory than RN50 due to its deeper architecture.

Other considerations include that models with higher VRAM requirements may not be suitable for deployment on devices with limited GPU memory, selecting a model involves balancing the need for speed, accuracy, and memory consumption, and increasing the batch size will proportionally increase VRAM usage. Careful management of batch sizes is necessary to avoid memory overflow.

### E. Summary and Recommendations

TABLE VI shows the pros, cons, and recommendations for the different CLIP models. To select a model between them, it is important to consider several factors. The first one is resource availability, which includes ensuring that computational resources (e.g., GPU memory, processing power) are sufficient for the chosen model, especially for larger models like ViT-L/14 and assessing VRAM availability to prevent out-of-memory errors during embedding. Secondly, application requirements, in which determining whether speed, accuracy, or memory efficiency is the priority and understanding that real-time applications may benefit from faster models with lower VRAM requirements, like ViT-B/32 or RN50. Finally, model familiarity, in which it is recommended to select a model architecture that aligns with the development team's expertise, while understanding that ResNet models may be preferable for those more familiar with convolutional neural networks and ViT models may offer advantages in tasks where capturing global context is important.

Nonetheless, selecting the appropriate CLIP model involves balancing the application's performance requirements with the available computational resources, including VRAM. CLIP ViT-B/32 and CLIP RN50 are recommended for applications prioritizing speed and memory efficiency. CLIP ViT-B/16 and CLIP RN101 offer a middle ground between performance, resource demands, and VRAM usage. For tasks where maximum accuracy and detail are critical, CLIP ViT-L/14 is ideal, provided that sufficient computational resources and GPU memory are available.

TABLE VI
PROS, CONS, AND RECOMMENDATIONS FOR DIFFERENT CLIP MODELS

| Model | Pros | Cons | Recommendations |
|---|---|---|---|
| CLIP ViT-B/32 | • Lowest VRAM requirement among ViT models (1–2 GB).<br>• Fastest processing speed among ViT models.<br>• Lower computational requirements. | • Larger patch size may miss fine-grained details. | • Suitable for applications where speed and low memory consumption are critical.<br>• Ideal for deployment on devices with limited GPU memory |
| CLIP ViT-B/16 | • Good balance between speed and performance. | • Higher VRAM requirement (2–3 GB) than ViT-B/32. | • Ideal for applications requiring a compromise between efficiency, |
| CLIP ViT-L/14 | - Better at capturing details than ViT-B/32.<br>- Highest performance in terms of detail and accuracy.<br>- Captures fine-grained features effectively. | - Slower than ViT-B/32.<br>- Highest VRAM requirement (4–6 GB).<br>- Slowest processing speed.<br>- Requires significant computational resources. | accuracy, and moderate memory usage.<br>- Best suited for applications where accuracy is paramount.<br>- Recommended only if ample computational resources and GPU memory are available. |
| CLIP RN50 | - Moderate VRAM requirement (2–3 GB).<br>- Efficient and fast due to convolutional operations.<br>- Well-established architecture with extensive community support. | - May not capture global relationships as effectively as ViT models. | - Appropriate for real-time applications with limited computational resources.<br>- Suitable when a balance between performance and memory consumption is needed. |
| CLIP RN101 | - Deeper network captures more complex features than RN50. | - Higher VRAM requirement (3–4 GB) than RN50.<br>- Slower than RN50.<br>- Increased computational cost. | - Suitable when a performance improvement over RN50 is desired.<br>- Consider VRAM limitations when deploying on hardware with restricted GPU memory. |

## IV. PROPOSED METHODOLOGY

Our methodology aims to establish a real-time scene understanding framework that serves as a critical component for advanced driver assistance systems. These systems necessitate rapid and accurate comprehension of driving environments to provide drivers with timely, context-aware advice. Our framework is built on two key pillars:

### A. Embedding Scene Images with Clip

We leverage the Contrastive Language-Image Pre-Training (CLIP) model to embed scene images into a high-dimensional vector space. CLIP is a robust model known for its ability to associate textual descriptions with visual data effectively. It captures semantic relationships between visual elements and their corresponding textual attributes, transforming complex visual scenes into embeddings suitable for computational processing. These embeddings allow our framework to map driving scenes into a representation that preserves the underlying semantic structure, making it ideal for real-time retrieval and analysis.

### B. Efficient Indexing with FAISS

To manage and search through the high-dimensional embeddings generated by CLIP, we incorporate Facebook AI Similarity Search (FAISS). FAISS is an advanced library optimized for similarity search and clustering of dense vectors. Known for its scalability and speed, FAISS is designed to handle large datasets with high-dimensional data efficiently. By utilizing FAISS, we index the CLIP-generated embeddings, enabling rapid retrieval of the most relevant scenes. This integration is crucial for real-time applications where speed and precision are essential, as FAISS supports approximate nearest neighbor (ANN) search, ensuring quick access to relevant information even in large-scale scenarios.

The integration of CLIP and FAISS creates a powerful real-time scene understanding framework. CLIP converts visual data into a searchable vector space, while FAISS ensures that the retrieval process is both swift and accurate. Together, they provide the computational backbone for processing and interpreting driving scenes promptly, empowering the system to deliver timely and precise advice to drivers, enhancing safety and decision-making.

### C. Inference Phase

During the inference phase, the framework processes a test scene through the CLIP model to generate its embedding. This embedding is then queried in the FAISS index to identify the nearest neighbor scenes. The textual attributes associated with these nearest neighbors are retrieved, and for each attribute, the system predicts the value based on the majority vote among the retrieved neighbors. This process ensures that the predictions are both data-driven and contextually relevant.

### D. Model Evaluation and Fine-Tuning

Our methodology is designed to evaluate various CLIP model variants across different numbers of nearest neighbors (NNs) and compare their performance against the baseline model, GPT-4o, in a zero-shot, in-context learning setting. Models are ranked based on their precision and recall performance across multiple attributes.

Following the evaluation, the top-performing CLIP models are selected for fine-tuning. Fine-tuning aligns the embedding space more closely with the semantic requirements of scene understanding, allowing the model to capture subtle nuances in driving contexts. The fine-tuned models are then used to embed the training data into a newly refined vector space. A new FAISS index is built using these embeddings, enabling the retrieval of nearest neighbors for test scenes that have been embedded using the fine-tuned models. This alignment significantly improves the model's ability to predict scene attributes with greater precision and recall. The combination of CLIP and FAISS not only ensures real-time performance but also provides a flexible framework that can adapt to new scenarios through fine-tuning. By aligning the embedding space to the semantics of scene understanding, the methodology enhances the accuracy and robustness of predictions.

To address the limitation of the CLIP model's text encoder, which imposes a 77-token restriction, we implemented an efficient text description generation approach using the following method. This method encodes textual descriptions by coding attribute values using numbers then converting them into a concise string, joining them in a predefined order with commas to ensure the text remains compact and within the token limit. By focusing solely on the attribute values, the encoded descriptions effectively preserve the semantic relationships between similar images, ensuring they are clustered closely in the latent space. Meanwhile, the detailed textual descriptions of the attributes are stored separately in a file, indexed for retrieval through FAISS. This dual approach guarantees computational efficiency while maintaining the richness of the original attribute descriptions for downstream tasks.

### E. Prompt Engineering

The prompt is designed for GPT-4o to analyze driving-related images from the perspective of an ego vehicle and classify detected features into predefined categories, as shown in FIGURE 1. It provides a structured framework for scene understanding by assigning values to categories such as road attributes, weather conditions, and complex road structures. However, compared to the CLIP retrieval system, this approach removes many attributes because the indexed images in this analysis represent only a single level or class. Additionally, temporal features like "Stages" (e.g., Approaching, Entering, Passing) are converted into a binary format for simplicity. For instance, categories such as "Rail Crossing" are represented as '0' if not detected and '1' if detected, effectively merging the temporal stages into a single binary outcome. This conversion aligns with the frame-based nature of the analysis in this paper, focusing on frame-level detection and classification rather than temporal sequences. This streamlined representation ensures the framework is optimized for frame-based image retrieval and analysis, critical for real-time and retrospective applications.

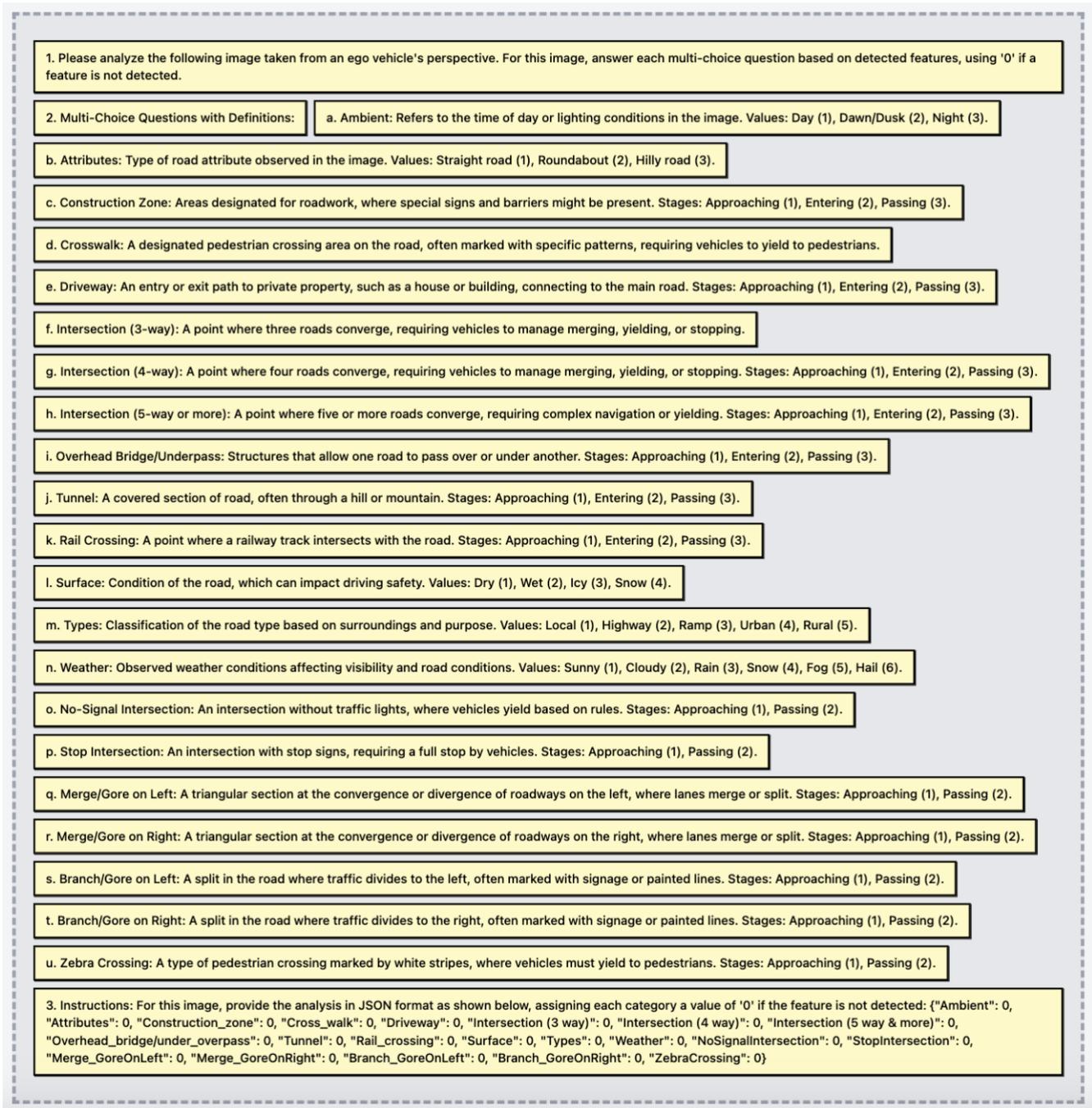

**Fig. 1.** Structured prompt for scene understanding by assigning values to categories such as road attributes, weather conditions, and complex road structures. The prompt shows all the attributes that were used to test the models.

## V. Dataset Description

The Honda Scenes Dataset (HSD) is a large-scale annotated dataset designed to support dynamic traffic scene classification. Comprising 80 hours of high-quality driving video data collected from the San Francisco Bay Area, HSD offers a rich variety of scenarios for training and testing. The dataset includes 11 road place classes, such as 3-way intersections, 4-way intersections, zebra crossings, and construction zones. It also spans four road environments (rural, urban, highway, and ramp) and four weather conditions (rainy, sunny, cloudy, and foggy). Detailed annotations are available in JSON and CSV formats, capturing temporal and frame-level attributes like road conditions, surface conditions, and event sub-classes such as "approaching," "entering," and "passing." Specialized sub-classes for merges and branches allow for granular analysis, enhancing the dataset's utility in various traffic-related research contexts.

The preprocessing methodology involves systematic steps to prepare the data for analysis, as shown in FIGURE 2. The dataset is pre-split into training and testing sets to ensure no spatial overlap between trips, maintaining the independence of test data for robust evaluation. Training data is outlined in trainsplit.txt, while testing data is defined in valsplit.txt. Videos are converted into individual frames to facilitate frame-level analysis. Frames in the test set are downsampled to one frame per second, while those in the training set are downsampled to one frame every two seconds, reducing computational demands while retaining essential information. Frames with all 21 attributes set to zero are filtered out, ensuring that only frames with meaningful annotations are included in the analysis. This preprocessing approach supports efficient and effective use of the dataset for training and evaluation purposes.

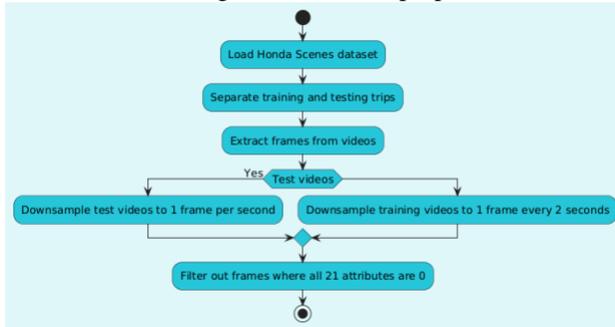

**Fig. 2.** Overview of the Data Preprocessing Workflow: Steps include splitting the Honda Scenes Dataset into training and testing, extracting frames from videos, downsampling at different rates for training and testing, and filtering out frames with no relevant information.

## VI. Experimental Work

### A. Pre-Trained Models Results

In this section, we address the first research question: how well do pre-trained models perform in the context of scene understanding? Our primary focus is to identify which of these relatively lightweight models, capable of real-time inference, achieve higher rankings compared to other CLIP models studied here and GPT-4o in zero-shot settings. To this end, we evaluated the performance of these models in terms of precision and recall across different class levels, aiming to pinpoint their strengths and limitations. Based on that, we ranked the models using the Euclidean distance between their positions in the precision-recall space and the optimal (1,1) point, providing a clear metric for performance comparison. This distance serves as a metric for ranking the models, with smaller distances indicating better performance. For visualization, we present a heatmap where each cell represents the distance of a model for a specific attribute-class combination. It is important to note that the CLIP models used in this study rely on retrieval-based mechanisms, where the number of nearest neighbors (indicated in parentheses) is adjusted for each model. Once the best-performing CLIP models are identified, we proceed to the fine-tuning phase for further performance enhancement. These results are presented in Fig. 3.

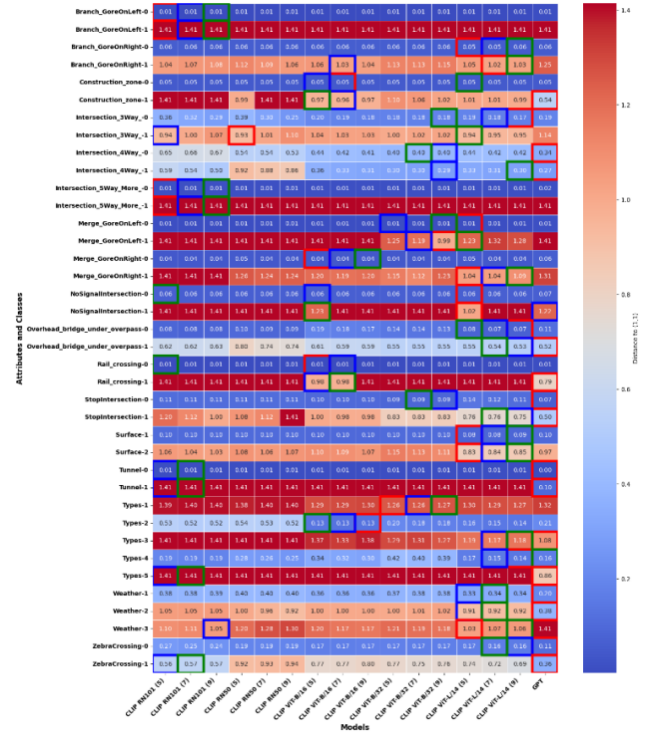

**Fig. 3.** Heatmap showing the pre-trained model rankings based on Euclidean distance to the ideal precision-recall point (1,1) for all attribute-class combinations, with top-performing models highlighted. The models with the top three performances (smallest distances) for each attribute-class combination are highlighted with bounding boxes of different colors: green for the best, blue for the second-best, and red for the third-best.

Fig. 3 shows that GPT-4o model consistently ranks as one of the top-performing models across multiple attribute-class combinations, indicated by its frequent appearance with minimal distances. Similarly, CLIP ViT-B/32 (5) and CLIP ViT-L/14 (5) models also demonstrate strong performance in several cases, especially for attributes like "Merge_GoreOnLeft" and "ZebraCrossing."

GPT-4o, as a zero-shot model, performs well in specific attributes, such as detecting the absence of construction zones, where it achieves balanced precision and recall. Meanwhile, the RN models (RN50 and RN101) exhibit lower overall performance compared to the CLIP models. These models tend to prioritize precision over recall, particularly in intersection scenarios, reflecting a conservative prediction strategy but with reduced sensitivity [27], [28].

Attribute-specific trends further illustrate these differences. For Branch and Merge Gore scenarios, CLIP models outperform GPT-4o, maintaining moderate precision and recall, while GPT-4o shows substantial limitations in generalizing to complex spatial relationships. In the Construction Zone attribute, GPT-4o excels in detecting non-construction zones but performs inconsistently in construction-specific scenarios. Conversely, CLIP models show a greater deviation from the optimal precision-recall point in Class 1. For intersections, as the complexity increases from 3-way to 5-way scenarios, all models struggle, with ViT-L/14 achieving the best precision but sacrificing recall. In the Overhead Bridge/Underpass attribute, Class 0 scenarios see all models performing well, but for Class 1, ViT-L/14 strikes a balance between precision and recall, outperforming GPT-4o, which shows high recall but lower precision.

A clear trend emerges across all attributes: models perform significantly better in simpler scenarios (Class 0) compared to complex ones (Class 1). The CLIP models consistently outperform GPT-4o and RN models in handling complex spatial reasoning tasks, emphasizing their robustness for such applications. Additionally, a trade-off between precision and recall is evident, with ViT-L/14 prioritizing precision while other models, such as ViT-B/16 and GPT-4o, emphasize recall.

The analysis highlights the strengths of CLIP models, particularly the ViT variants, in delivering robust performance for complex scene understanding. GPT-4o, while commendable in simpler scenarios, is less reliable in domain-specific tasks. The observed trade-offs between precision and recall suggest that hybrid approaches or fine-tuning could further improve model performance, especially in dynamic and nuanced driving contexts.

### B. Fine-Tuned Models Results

As CLIP ViT-B/32 (5) and CLIP ViT-L/14 (5) showed the best performance among all models, we chose to fine-tune them. The training data preparation involves pairing images with textual descriptions generated from associated attributes, extracted from CSV files. Each image is processed and matched with a descriptive text string representing its attributes to create input pairs for the model. The training process employs the Adam with Weight Decay optimizer with a learning rate of $10^{-5}$. The model is trained over 8 epochs using mixed-precision training to enhance computational efficiency. Performance is monitored by tracking loss values at both batch and epoch levels, ensuring consistent evaluation throughout the training.

The heatmap illustrates the distances between the precision-recall points of various models and the ideal point (1,1) in the precision-recall space, aggregated across attributes and scales. The performance of the models is represented with a color gradient, where lower distances (closer to blue) indicate better alignment with the ideal, and the top three performing models for each attribute-class pair are highlighted. The comparison of fine-tuned and non-fine-tuned CLIP ViT-B/32 (5) and CLIP ViT-L/14 (5) models, as well as GPT-4o in its zero-shot configuration, reveals key insights into model performance and adaptability.

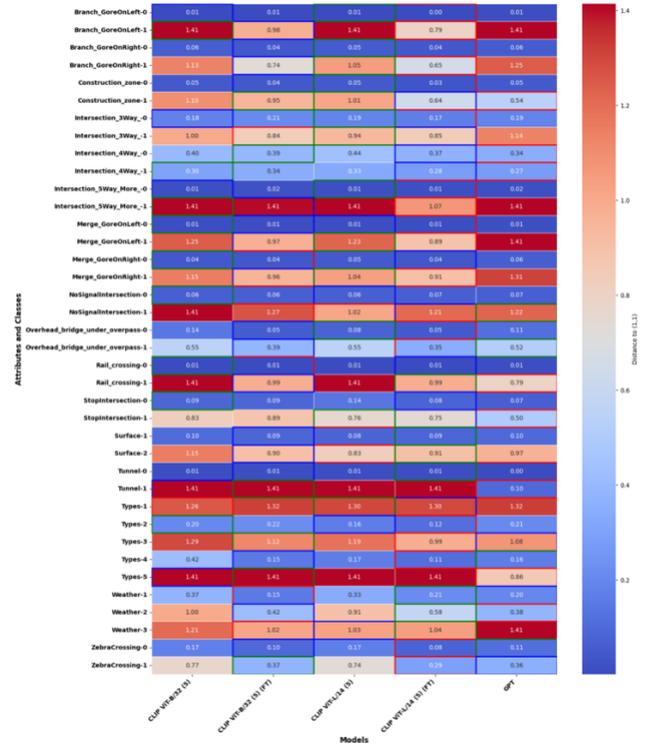

**Fig. 4.** Heatmap showing the performance improvement of fine-tuned models (FT) compared to non-fine-tuned counterparts and GPT-4o in zero-shot, highlighting top-performing models across attributes and scales. The models with the top three performances (smallest distances) for each attribute-class combination are highlighted with bounding boxes of different colors: green for the best, blue for the second-best, and red for the third-best.

The fine-tuned models of CLIP ViT-B/32 (5) and CLIP ViT-L/14 (5), which are labeled as FT in Fig. 4, demonstrate consistent improvements over their non-fine-tuned counterparts. This is evident in their reduced distances across multiple attribute-class pairs, such as ZebraCrossing-1, Weather-2, and StopIntersection-1. Fine-tuning allows these models to adapt effectively to the nuances of the dataset, enhancing their precision and recall and bringing their performance closer to the ideal. This adaptability highlights the importance of leveraging fine-tuning to optimize lightweight models for specific tasks.

When compared to GPT-4o in its zero-shot configuration, the fine-tuned models frequently outperform or match its performance. For instance, CLIP ViT-L/14 (5) (FT) surpasses GPT-4o in attributes such as Types-2 and Tunnel-1, while CLIP ViT-B/32 (5) (FT) achieves competitive results in attributes like Merge GoreOnLeft-1 and Surface-2. This is particularly significant given GPT-4o's reliance on large-scale pretraining, demonstrating that task-specific fine-tuning can enable lightweight models to rival or exceed the capabilities of more resource-intensive models.

## VII. Discussion

To make a direct comparison across all attributes, Fig. 5

illustrates the aggregated performance of CLIP ViT-B/32 (5) and CLIP ViT-L/14 (5) with their fine-tuned counterparts and GPT-4o in zero-shot settings, measured using the mean of weighted Precision, Recall, and F1-scores. Results show the Euclidean distance to the ideal Precision-Recall point (1,1), where smaller values indicate better performance. Fine-tuned models exhibit significantly lower distances, demonstrating their improved alignment with the ideal point. Results of F1-scores showed that CLIP ViT-L/14 (5) (FT) achieved the highest value of about 91.1% followed by CLIP ViT-B/32 (5) (FT) with 90.5%, highlighting their careful balance between Precision and Recall. These results emphasize the effectiveness of fine-tuning in enhancing model performance compared to zero-shot or pre-trained approaches.

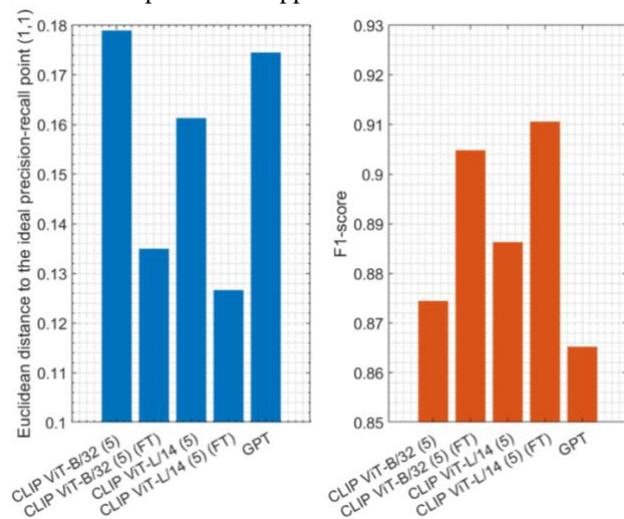

**Fig. 5.** Aggregated performance of CLIP ViT-B/32 (5) and CLIP ViT-L/14 (5) with their fine-tuned counterparts and GPT-4o in zero-shot settings, measured using the mean of weighted Precision, Recall, and F1-scores.

The results underscore the potential of lightweight models like CLIP ViT-B/32 and CLIP ViT-L/14 for real-time deployment on edge devices. These models are computationally efficient, making them suitable for dynamic, resource-constrained environments such as ADAS or other autonomous vehicles systems. Fine-tuning further enhances their suitability by aligning their performance with specific datasets, ensuring both accuracy and efficiency in real-time applications.

## VIII. CONCLUSIONS

This study presents a robust framework for real-time scene understanding in, leveraging the capabilities of the CLIP model, which can be used for real-time deployment on edge devices. A CLIP-based retrieval system was fine-tuned for real-time ADAS applications, enabling accurate classification of dynamic driving scenes. The model was benchmarked against state-of-the-art in-context learning methods, including GPT-4o in zero-shot scenarios, where it demonstrated superior performance, particularly in complex scenarios. Comprehensive frame-level evaluations were conducted using the Honda Scenes Dataset, which includes 80 hours of annotated driving videos capturing diverse road and weather conditions. These evaluations highlighted the robustness of the framework across varied environments. Additionally, the integration of language-based supervision with visual scene recognition enhanced the semantic understanding of traffic scenes, contributing valuable insights to road safety and autonomous driving research.

By embedding scene images into a high-dimensional vector space using CLIP and indexing them with FAISS, the framework enabled rapid and precise retrieval of relevant scenes. This feature is critical for real-time ADAS applications. Fine-tuning the CLIP models, such as ViT-L/14 and ViT-B/32, improved their performance by aligning embeddings with the semantic requirements of scene understanding. Notably, the fine-tuned ViT-L/14 model achieved the highest F1-score of 91.1%, followed closely by ViT-B/32 at 90.5%.

Precision and recall analysis revealed trade-offs between these metrics, with different models emphasizing either precision (e.g., ViT-L/14) or recall (e.g., ViT-B/16). This finding underscores the need to select models based on the specific requirements of a given application. The lightweight nature of models like ViT-B/32 and ViT-L/14, combined with their computational efficiency, positions them as ideal candidates for deployment on edge devices in resource-constrained environments such as autonomous vehicles.

As fine-tuning significantly enhanced the models' ability to classify complex scenes, this positions CLIP as a scalable, accurate, and efficient solution for dynamic scene understanding. Furthermore, the study's exploration of multimodal techniques, including advanced indexing and semantic-rich embeddings, provides a solid foundation for future advancements in AI-driven scene classification systems. Practically, this study establishes a pathway toward safer, smarter, and more context-aware autonomous driving systems. The integration of advanced techniques and models like CLIP ensures precise, real-time decision-making capabilities, enabling more effective ADAS and paving the way for further innovations in autonomous vehicle technology.